# DENet: A Universal Network for Counting Crowd with Varying Densities and Scales

Lei Liu, Jie Jiang, Wenjing Jia, *Member, IEEE*, Saeed Amirgholipour, Michelle Zeibots, Xiangjian He, *Senior Member, IEEE*,

*Abstract*—Counting people or objects with significantly varying scales and densities has attracted much interest from the research community and yet it remains an open problem. In this paper, we propose a simple but an efficient and effective network, named DENet, which is composed of two components, i.e., a detection network (DNet) and an encoder-decoder estimation network (ENet). We first run DNet on an input image to detect and count individuals who can be segmented clearly. Then, ENet is utilized to estimate the density maps of the remaining areas, where the numbers of individuals cannot be detected. We propose a modified Xception as an encoder for feature extraction and a combination of dilated convolution and transposed convolution as a decoder. In the ShanghaiTech Part A, UCF and WorldExpo'10 datasets, our DENet achieves lower Mean Absolute Error (MAE) than those of the state-of-the-art methods.

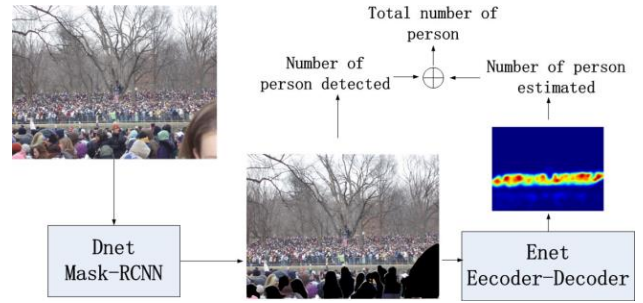

Fig. 1. The architecture of the DENet

## I. INTRODUCTION

Vision-based techniques for accurately counting or estimating the number of people (or objects) in a crowded scene are desirable techniques in many real world applications including visual surveillance, traffic monitoring and crowd analysis. This is true especially in restrictedly, public places such as train stations, where incidents, traffic delay and even terrible stampedes have been reported due to overcrowding in these places. However, various real-world situations, such as heavy occlusions, cluttered background, size and shape variations of people, and perspective distortion, have posed great challenges for practical solutions capable of handling such situations. Thus, correctly counting in crowded scenes is still an open and popular research problem nowadays.

Early works addressing the crowd counting problem mainly follow the counting-by-detection framework [1], which uses body or part-based detectors to detect individual people in crowd images. These methods require well-trained classifiers to extract low-level features (e.g., Haar wavelets [2] and HOG-histogram oriented gradients [3]) from a whole human body. Recent approaches seeking an end-to-end solution using CNN-based object detectors such as YOLO3 [4], SSD [5], Fast R-CNN and Faster R-CNN [6] have greatly improved the counting accuracy. Mask R-CNN [7], proposed by He, can not only detect objects, but also segment them from background. Although detection-based crowd counting methods are successful for dealing with scenes with low crowd density, when it comes to highly congested environments, where only parts of the whole objects are visible, the performance of these detection-based approaches, affected by the size of target and occlusions, always degrades significantly, posing great challenges to object detectors.

The feature-regression-based approaches, as shown in [8], [9], [10], [11] etc., aim to obtain the density function of an image containing people and then calculate the total count by integrating the density function over the whole image space. More features, such as foreground and texture features, have been used for generating low-level information [12]. Following similar approaches, Idrees et al. [13] proposed a model to extract features by employing Fourier analysis and SIFT (Scale-Invariant Feature Transform) [14] interest-point based counting. They have demonstrated a countable solution for handling highly crowded scenes.

Recently, CNN-based approaches have shown a remarkable success for crowd counting due to their excellent representation learning ability. Zhang et al. [8] designed a multi-column CNN (MCNN) to tackle the large scale variation in crowd scenes. With a similar idea, Onoro and Sastre [10] proposed a scale-aware network, called Hydra, to extract features at different scales. Very recently, inspired by MCNN, Sam et al. [9] presented the Switch-CNN, which trained a classifier to select the optimal regressor from multiple independent regressors for specific input patches. Zhang et al.[15] proposed a baseline of crowd segmentation and estimation. Sindagi et al. [16] proposed to consider the global and local contextual information by using four modules. They used a combination of adversarial loss and pixel-wise Euclidean loss to improve the density map accuracy.

Lei Liu and Jie Jiang are in School of Instrumentation Science and Opto-Electronics Engineering, Beihang University, China. e-mail:by1417114@buaa.edu.cn; jiangjie@buaa.edu.cn

Sean He, Wenjing Jia and Saeed Amirgholipour are in Global Big Data Technologies Centre, University of Technology Sydney, Australia. Michelle Zeibots is in Institute for Sustainable Futures, University of Technology Sydney, Australia. email:Xiangjian.He@uts.edu.au;Wenjing.Jia@uts.edu.au; Saeed.AmirgholipourKasmani@student.uts.edu.au;Michelle.E.Zeibots@uts.edu.au

Corresponding author:Sean He and Jie Jiang

Liu et al.proposed the DecideNet [17], which adaptively adopted detection and regression based count estimations under the guidance from an attention mechanism. Li et al. [18] proposed the CSRNet by using VGG-16 [19] to extract feature and dilated convolution layers to generate density map. Cao proposed the SANet [20] by combining the Euclidean loss and counting loss, and using a set of transposed convolutions to create high-resolution density maps.

The most recent works, e.g., [8] [9] [16], have attempted to address the scale variation issue with multi-scale architectures. They used CNNs with different field sizes to extract features adaptive to the large variation in people sizes and achieved significant improvements. Since the high-resolution density maps contain finer details, we believe that it is of great value to develop crowd density estimation techniques that can produce high-resolution and high-quality density maps. However, there exist the following limitations in the existing crowd counting works dealing with varying crowdedness. Most of the works have focused on either density estimation or people detection. Although some of the recent works (e.g., DecideNet [17]) have attempted to develop an adaptive network by combining density estimation and people detection, they have very inflexible solutions. In the existing CNN-based methods, the input of an estimation network is a whole image, where people of different scales are all represented as dots of the same scale in the heat map. It is difficult to learn a network to generate the same result with inputs of different scales.

Based on our observation, for objects with significantly varying scales (see Fig. 1), the areas where people can be clearly detected often degrade the network performance greatly. Why not to take the advantages of both counting-by-detection and counting-by-estimation? This is to detect clear persons in low density areas and to estimate people density in high density areas. Note that, similar ideas have been explored in references. For example, the DecideNet [17] introduces a ratio that needs to be retrained each time and is the ratio of high-density areas to the low-density areas changes. This inflexibility poses lots of practical issues for application. In our work, we propose a simple but an efficient and effective solution. We first run a detection network (the 'DNet') on an input image to detect and count the detected people who can be segmented clearly. These segmented people areas are then removed from the input image. Then, an encoder-decoder estimation network (the 'ENet') is utilized to generate the density map over the crowd areas where individual people cannot be separated. Furthermore, to combine the advantages of detection and estimation, we propose a novel loss function to train the ENet. The loss is composed of an Euclidean loss and a counting loss. The Euclidean loss is used to generate accurate heat maps and the counting loss is to compare the number of predicted people and the ground truth.

The main contributions of our work are summarized as follows.
- We propose a novel network structure, namely Detection-Estimation Network (simplified as the 'DENet') for accurately and efficiently counting crowds of varying densities. The structure improves the multi-scale representation of the learned network and can produce high-resolution density maps. When applying our DENet struture to some state-of-the-art crowd-counting networks, all estimation accuracies can be improved to some extent, demonstrating the applicability of our core idea.
- To further improve the generality of the estimation network for different scales, we propose a novel estimation network (the 'ENet'), which uses a modified Xception architecture as the encoder and combines the dilated convolution and transposed convolution as its decoder.
- We propose a new loss function for training two networks jointly. The function combines an Euclidean loss and a newly proposed counting loss. The Euclidean loss measures the estimation error at pixel level and the counting loss measures the counting error of people over the whole image.
- Extensive experiments on several challenging benchmarks are conducted to demonstrate the superior performance of our approach over the state-of-the-art solutions.

## II. PROPOSED ALGORITHM

To address the varying scale issues, we follow the two points discussed above and propose a novel encoder-decoder network, named Detection-Estimation Network (DENet), of which architecture is shown in Fig. 1. Inspired by the success of Mask R-CNN [7] on object detection, we first adopt Mask R-CNN to detect and segment people who can be clearly differentiated from crowd, and then we propose a novel estimation network to estimate the density map for the areas where individuals cannot be segmented due to high crowdedness. We modify the Xception [21] to be the encoder of the estimation network so as to improve the representation ability and scale diversity of features. The decoder is composed of a set of dilated convolutions [22] and transposed convolutions. It is used to generate high-resolution and high-quality density maps, of which the sizes can be exactly the same as that of the input images. This section presents the details of our proposed DENet. Moreover, we propose a new loss function.

### A. DNet

The Mask R-CNN proposed in [7], extends Faster R-CNN [6] by adding a branch for predicting an object mask in parallel with the existing branch for bounding box recognition. Thus, Mask R-CNN can not only detect objects but also segment them from the input image. We adopt Mask R-CNN as our DNet in our work and we only retain large people segmented by DNet, due to the fact that the smaller the size of people is, the more miss or false detection the counting has.

### B. ENet

As shown in Fig. 1, our estimation network consists of two components, i.e., feature map encoder (FME) and density map estimator (DME). We adopt a modified Xception as the FME to extract features, and a set of dilated convolutions and transposed convolutions as DME to create high-resolution and high-quality density maps.

FME: Following the similar idea in [21], we modify the Xception to form an FME of the estimation net because of it has been widely used as the encoder for feature extration. We find that by removing some blocks from the original Xception architecture, its accuracy will not be affected. In order to reduce the computation complexity, The architecture of the modified Xception is shown in Fig. 2.

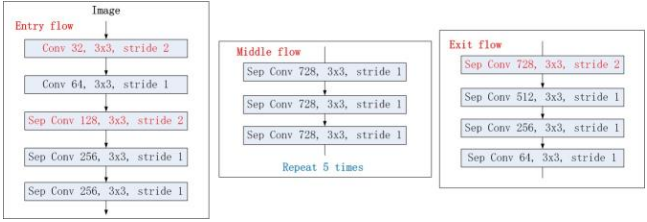

Fig. 2. The modified Xception in our DNet

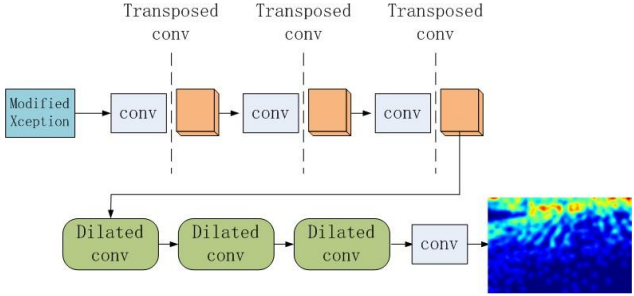

Fig. 3. The architecture of the DME in our DNet

For the DME, while density-estimation-based approaches take the spatial information into account, the outputs of these works are mostly of low-resolution due to several pooling layers, and hence cause the loss of details compare with the groundtruth. Inspired by the approaches of CSRNet [18] and SANet [20], the dilated convolution can keep more details than the traditional convolution and the transposed convolution can alleviate the loss of information. Therefore, in our work, we deploy a combination of dilated convolutions and transposed convolutions as the decoder for the ENet to create high-resolution and high-quality density maps.

Fig. 3 shows the architecture of the DME of our ENet. Three pairs of convolution layer and transposed convolutional layer are added after FME, the size of the filters in convolution layer ranging from $7 \times 7$ to $3 \times 3$, and each transposed convolutional layer double the size of the output of previous layer. A set of dilated convolutional layers are added after the last transposed convolutional layer to keep the details of feature maps. Then, a $1 \times 1$ convolution layer is added after the last dilated convolutional layer to generate density maps. ENet focuses on the features of high density part in the input image, and hence facilitates the feature learning in the model. The size of density maps generated by ENet are the same as the size of input image.

### C. Loss function

In our DENet, to create an accurate density map, we propose to use the Euclidean loss and a counting loss. The Euclidean loss is used to measure the density map error at pixel level. We propose a counting loss to measure the counting error, which is the difference between the ground-truth number of people and the sum of the number of people obtained by detection and the number of people obtained by density estimation.

Let $L_E$ denote the Euclidean loss, which measures the estimation error at pixel level and is defined by:

$$L_E = \frac{1}{N} ||F(X) - Y(X)||_2^2, \quad (1)$$

where $N$ is the number of pixels in the density map, $X$ is the input image, $Y(X)$ is the corresponding ground-truth density map, and $F(X)$ denotes the estimated density map. We use $L_C$ to measure the difference between the ground truth number of people and the sum of the numbers of people detected and estimated. $L_C$, which representing the computing loss, is defined by:

$$L_C = ||\frac{(N_{GT} - N_D - N_E)}{(N_{GT} - N_D + 1)}||_2^2, \quad (2)$$

where $N_{GT}$ is the ground-truth of the total number of people in an input image, and $N_D$ and $N_E$ are the numbers of people obtained by the detection and estimation, respectively. In case DNet detects all the people, we add 1 in the denominator of this equation so that the demoninator is never be zero. By weighting the above two loss functions, we define the final loss function as:

$$Loss = L_E + \alpha L_C, \quad (3)$$

where $\alpha$ is the weight to balance the Euclidean loss and counting loss. In our experiments, we empirically set $\alpha$ as 0.1.

## III. IMPLEMENTATION DETAILS

In this section, we provide more technical details of training the proposed DENet, generating ground truth and performance evaluation metrics.

### A. DENet training

The weights in DNet are from a well-trained Mask R-CNN. For the ENet, all of the conv layers are initialized with zero. Training strategy of ENet Adam and the learning rate is decrease with the iteration increase. In the training stage, we use the whole image to train the network, and each training image is used four times to augument data.

### B. Ground truth generation

Ground truth generation is similar with those in the existing works [8], [18], [20], annotations for crowd images are dots at the center of pedestrians' heads, and the ground truth density functions are generated at each of the dots. We generate the ground truth by bulrring the dot map with Gaussian kernel(which is normalized to 1). The density map is defined by:

$$F(x) = \sum_{i=1}^{} \sigma(x - x_i) * tt_{\sigma_i}(x), \quad (4)$$

To generate the density map, the dot map $\sigma(x - x_i)$ is convolved with a Gaussian kernel with a standard deviation

$\sigma_i$, where $x_i$ is the position of pixel of the $i^{th}$ dot in the image. In our experiments, we follow the configuration in [18], where different datasets corresponding to different $\sigma$ of Gaussian kernel.

Note that all the Gaussian functions are summed and normalized, so the total object count is preserved even when there is overlapping between targets.

*C. Evaluation metrics*

For crowd density estimation, two metrics are commonly used for quantitatively measuring the counting error in previous works, i.e., Mean Absolute Error (MAE) and Mean Squared Error (MSE), defined by:

$$MAE = \frac{1}{N}\sum_{i=1}^{N} |C_i - C_i^{GT}| \quad (5)$$

and

$$MSE = \sqrt{\frac{1}{N}\sum_{i=1}^{N} |C_i - C_i^{GT}|^2}, \quad (6)$$

where $N$ is the number of images in test set and $C^{GT}$ is the ground truth of number of people in the $i^{th}$ test image. $C_i$ is the number of estimated counting in the $i^{th}$ test image, which is defined by:

$$C_i = N_E + N_D, \quad (7)$$

$N_D$ and $N_E$ are the numbers of people obtained by the detection and estimation, respectively. Roughly speaking,Roughly speaking, the lower the MAE and MSE is, the better accuracy the estimation method has.

## IV. EXPERIMENTS

We evaluate the performance of our approach on four different public datasets, i.e., ShanghaiTech, UCF, UCSD, and WorldExpo'10 [13], [23], [8], [24] and compare with the state-of-the-art methods [18], [20]. In this section, we first present the comparative experimental results on the above benchmark datasets, and then an ablation study conducted on the ShanghaiTech dataset is included to analyze our baseline idea of combining detection and estimation. The implementation of our approach is based on the PyTorch framework [25].

*A. Results on ShanghaiTech dataset*

ShanghaiTech crowd counting dataset was firstly introduced by Zhang et al. [8], 1198 annotated images are included and it has a total of 330,165 annotated person. This dataset is composed of two parts, where 482 images in Part A and 716 images in Part B, the data in Part A are downloaded from internet and the data in Part B are captured from streets in Shanghai, and Part A has more person in an image than Part B. Part A has 300 images for training and 182 images for testing, and Part B has 400 images for training and 316 images for testing. The result of our method and other recent works are shown in Table I.

As shown, our method achieves the lowest MAE (the highest accuracy) in ShanghaiTech Part A and also achieves second lowest MAE on ShanghaiTech Part B.

TABLE I
COMPARISON WITH STATE-OF-THE-ART METHODS ON THE SHANGHAITECH DATASET

| Method | partA | | partB | |
|---|---|---|---|---|
| | MAE | MSE | MAE | MSE |
| Zhang et al.[24] | 181.8 | 277.7 | 32.0 | 49.8 |
| MCNN [8] | 110.2 | 173.2 | 26.4 | 41.3 |
| Huang et al [26] | - | - | 20.2 | 35.6 |
| Switch-CNN[9] | 90.4 | 135.0 | 21.6 | 30.1 |
| CP-CNN [16] | 73.6 | 106.4 | 20.1 | 30.1 |
| CSRNet[18] | 68.2 | 115.0 | 10.6 | 16.0 |
| SANet[20] | 67.0 | 104.5 | **8.4** | **13.6** |
| DENet (ours) | **65.5** | **101.2** | 9.6 | 15.4 |

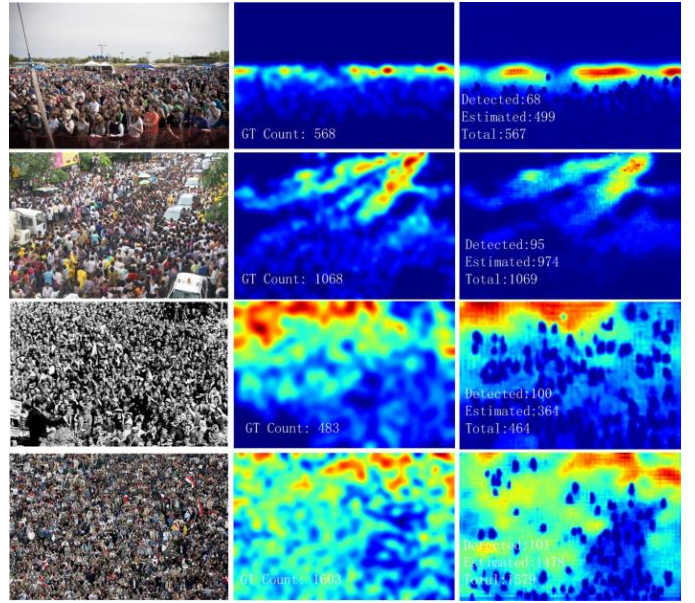

Fig. 4. Visualization of the estimated density maps of ShanghaiTech Part A using our approach.

Some sample images together with their results can be found in Fig. 4 on ShanghaiTech Part A and Fig. 5 on ShanghaiTech Part B.

*B. Results on UCF CC 50 dataset*

The UCF CC 50 dataset includes 50 annotated images with 63974 persons in total, it is downloaded from internet with different perspective and resolutions [13]. The number of annotated person in one image ranges from 94 to 4543, and the average number of persons per image is 1280. 5-fold cross-validation is used to evaluate the performance, which following the standard setting in [13]. The comparison results are shown in Table II and the visual qualities of generated density maps can be found in Fig. 6.

*C. Results on WorldExpo'10 dataset*

The WorldExpo'10 dataset [24] includes 3980 annotated images from 1132 video sequences that taken from 108 different surveillance cameras. This dataset consist 3380 images from training, and have 5 subset for testing with different scense, each testing subset has 120 images. We train our model following the instructions given in SectionIII-A. Results are shown in Table III and Fig. 7. The proposed DENet delivers the best MAE in Scene1, Scene2 and Scene3, and it achieves the best accuracy on average.

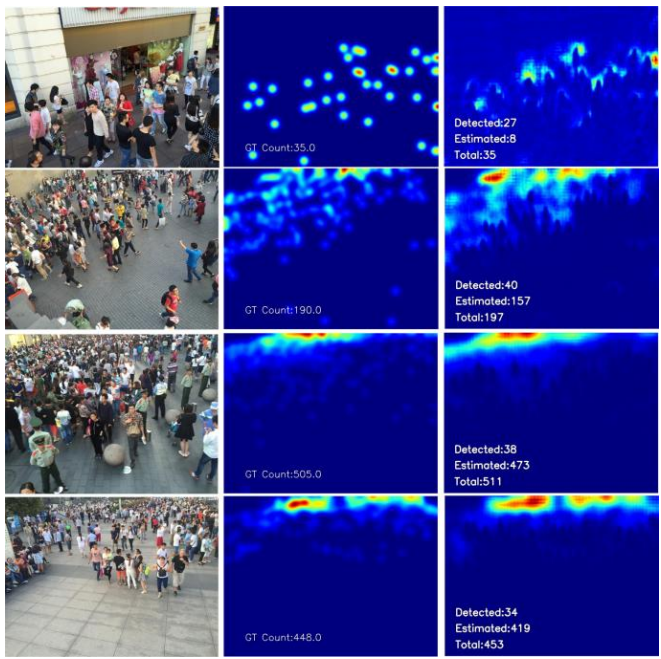

Fig. 5. Visualization of the estimated density maps of ShanghaiTech Part B for using our approach.

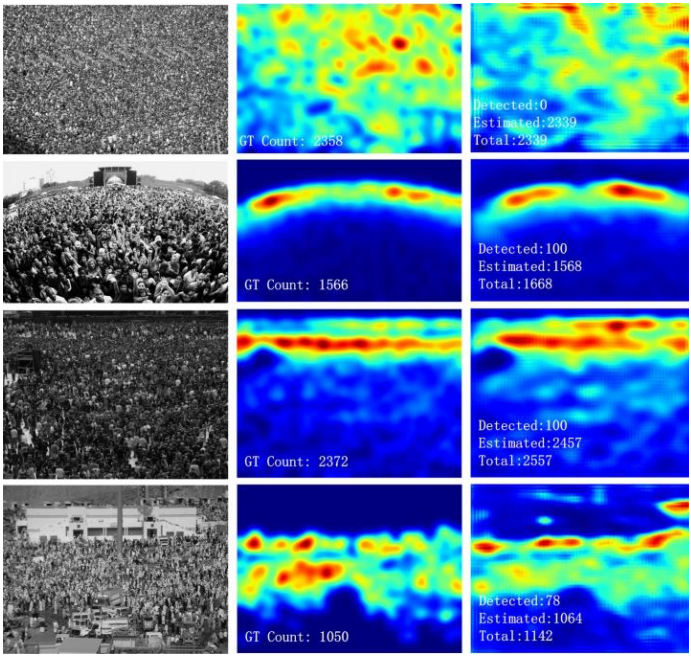

Fig. 6. Visualization of the estimated density maps of UCF dataset using our approach.

TABLE II
COMPARISON WITH STATE-OF-THE-ART METHODS ON THE UCF DATASET

| Method | MAE | MSE |
|---|---|---|
| Zhang et al.[24] | 467.0 | 498.5 |
| MCNN [8] | 377.6 | 509.1 |
| Huang et al [26] | 409.5 | 563.7 |
| Hydra-2s[10] | 333.7 | 425.3 |
| Switch-CNN[9] | 318.1 | 439.2 |
| CP-CNN [16] | 295.8 | 320.9 |
| CSRNet[18] | 266.1 | 397.5 |
| SANet[20] | 258.4 | **334.9** |
| DENet (ours) | **241.9** | 345.4 |

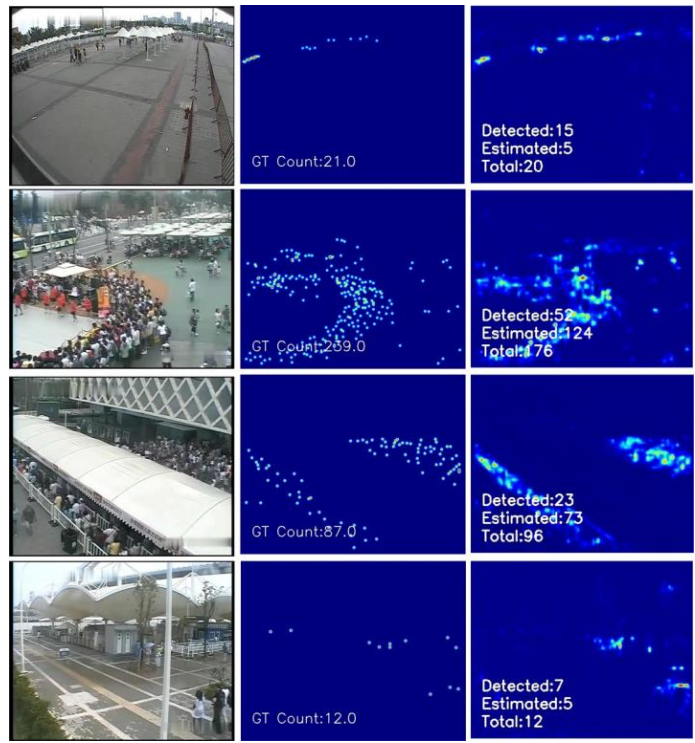

Fig. 7. Visualization of the estimated density maps of WorldExpo'10 for using our approach.

*D. Results on UCSD dataset*

The UCSD dataset [23] has 2000 images taken from surveillance cameras. The person in one image varying from 11 to and the size of each person in the image are similar. Following the [23], the training set contains 800 images and the rest of 1200 images are used for testing. Most people can be detected by DNet, and the results of UCSD dataset are shown in Table IV. The proposed algorithm outperforms the exisitng methods except SANet[20] and Huang et al [26] in the MAE category. We provide more results in Table IV and Fig. 8.

*E. Ablation experiments*

To demonstrate the effectiveness of our idea of DENet, we implement the MCNN and CSRNet to train them with our baseline idea of combing detection with the Mask R-CNN and their density-estimation networks. Based on the MCNN and CSRNet models, several ablation studies are conducted on the ShanghaiTech Part A and ShanghaiTech Part B dataset. The evaluation results are reported in Table V. As shown in this table, all results are improved over those originally reported in [8] and [18]. Thus, we can make a conclusion that most of the crowd counting estimation networks can be improved with our baseline idea.

TABLE III
COMPARISON WITH STATE-OF-THE-ART METHODS ON WORLDEXPO'10 DATASET

| Method | Scene1 | Scene2 | Scene3 | Scene4 | Scene5 | Average |
|---|---|---|---|---|---|---|
| Zhang et al.[24] | 9.8 | 14.1 | 14.3 | 22.2 | 3.7 | 12.9 |
| MCNN [8] | 3.4 | 20.6 | 12.9 | 13.0 | 8.1 | 11.6 |
| Huang et al [26] | 4.1 | 21.7 | 11.9 | 11.0 | 3.5 | 10.5 |
| Switch-CNN[9] | 4.4 | 15.7 | 10.0 | 10.4 | 5.8 | 8.9 |
| CP-CNN [16] | 2.9 | 14.7 | 10.5 | 10.4 | 5.8 | 8.9 |
| CSRNet[18] | 2.9 | 11.5 | 8.6 | 16.6 | **3.4** | 8.6 |
| SANet[20] | 2.8 | 14.0 | 10.2 | **12.5** | 3.5 | 8.6 |
| DENet(ours) | **2.8** | **10.7** | **8.6** | 15.2 | 3.5 | **8.2** |

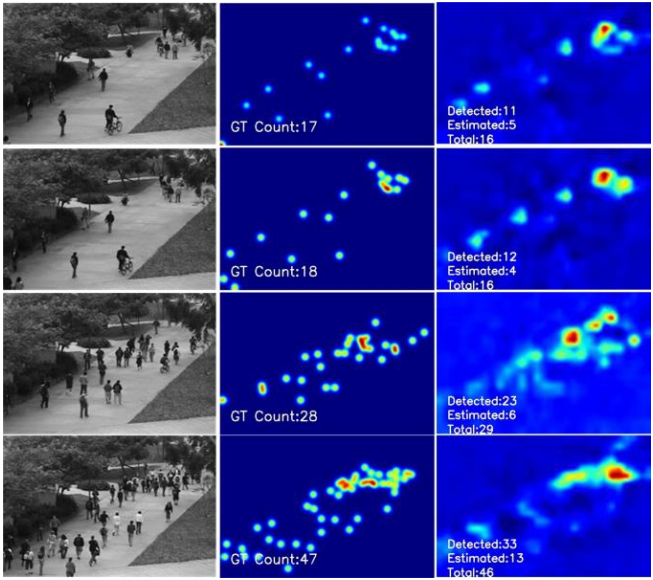

Fig. 8. Visualization of the estimated density maps of UCSD for using our approach.

TABLE IV
COMPARISON WITH STATE-OF-THE-ART METHODS ON THE UCSD DATASET

| Method | MAE | MSE |
|---|---|---|
| Zhang et al.[24] | 1.60 | 3.31 |
| MCNN [8] | 1.07 | 1.35 |
| Huang et al [26] | **1.00** | 1.40 |
| CCNN[10] | 1.51 | - |
| Switch-CNN[9] | 1.62 | 2.10 |
| CSRNet[18] | 1.16 | 1.47 |
| SANet[20] | 1.02 | **1.29** |
| DENet (ours) | 1.05 | 1.31 |

TABLE V
COMPARISON OF THE ESTIMATION ERROR OF TWO DIFFERENT NETWORK CONFIGURATIONS, I.E., MCNN [8] AND CSRNET [18], COMBINING WITH OUR IDEA, I.E., MASK R-CNN + MCNN AND MASK R-CNN + CSRN$_{ET}$.

| | partA | | partB | |
|---|---|---|---|---|
| Method | MAE | MSE | MAE | MSE |
| MCNN [8] | 110.2 | 173.2 | 26.4 | 41.3 |
| Mask R-CNN + MCNN | **105.6** | **164.1** | **23.2** | **37.5** |
| CSRNet [18] | 68.2 | 115.0 | 10.6 | 16.0 |
| Mask R-CNN + CSRNet | **67.5** | **112.1** | **10.1** | **15.5** |

## V. CONCLUSION

In this paper, we proposed a novel encoder-decoder architecture called DENet for accurate crowd counting. We used the Mask R-CNN to detect and segment clear persons, and have proposed a novel estimation network to create density maps. The numbers of detection and estimation are added to calculate the total number of persons. By taking advantage of the dilated convolutional layers and transposed layer, our estimation network can create high-quality density maps without losing resolution. We have proposed a new loss function that combines counting loss and Euclidean loss to train our estimation network. Experiments have shown that our method achieves the better performance on some major crowd counting datasets over the state-of-the-art methods.

## ACKNOWLEDGMENT

This work was partly supported by National Natural Science Fund of China (No. 61725501), Specialized Research Fund for the Doctoral Program of Higher Education of China (No. 20121102110032), and Rail Manufacturing CRC and Sydney Trains with UTS project ID PRO17-3968.

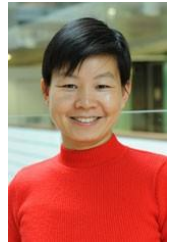

**Wenjing Jia** received her PhD degree in Computing Sciences from UTS in 2007. She is currently a Senior Lecturer at the Faculty of Engineering and Information Technology (FEIT), University of Technology Sydney (UTS). Her research falls in the fields of image processing and analysis, computer vision and pattern recognition.

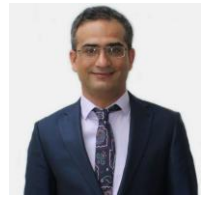

**Saeed Amirgholipour** received his Master degree in Computing Sciences from Isfahan university in 2009. Between 2011 to 2017 He was lecturer at Azad University, now he is PhD student in University of Technology, Sydney, Australia. His research interest include computer vision, deep learning video analytics, and image Sentiment analysis.

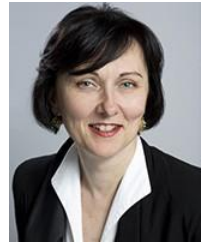

**Michelle Zeibots** is a transport planner, specialising in the analysis of sustainable urban passenger transport systems. Her research, consultancy work and teaching draws together operational, behavioural and governance features relating to multi-modal urban transport networks.

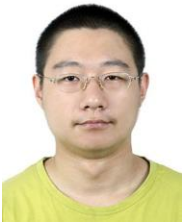

**Lei Liu** received the B.Sc and M.Sc Degrees in University of Science and Technology Beijing (USTB) from 2006 to 2014. He is currently a PhD student in the School of Instrumentation Science and Opto-electronics Engineering, Beihang University (BUAA), China. He was a visiting student in University of Technology Sydney from 2017 to 2019. His research interests is deep learning and computer vision.

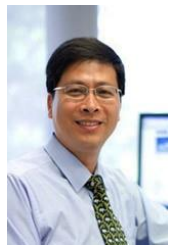

**Sean He** received his PhD degree in University of Technology, Sydney, Australia, in 1999. Since 1999, he has been with the Unniversity of Technology, Sydney, Australia. His research interests include image processing, network security, pattern recognition, computer vision and machine learning.

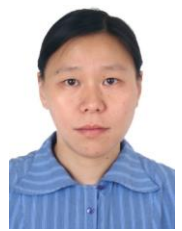

**Jie Jiang** received her Bachelors, Masters, and Ph.D. degrees from Tianjing University from 1991 to 2000. She is currently a Professor of the School of Instrumentation Science and Opto-electronics Engineering, Beihang University (BUAA), China. She has authored more than 50 articles and 30 inventions. Her research interests include image processing and machine vision.